\documentclass{article}

\usepackage[preprint]{corl_2026} % Uncomment for pre-prints (e.g., arxiv); This is like ``final'', but will remove the CORL footnote.

\usepackage{graphics} % for pdf, bitmapped graphics files
\usepackage{epsfig} % for postscript graphics files
\usepackage{amsmath} % assumes amsmath package installed
\usepackage{amssymb}  % assumes amsmath package installed
\usepackage{xcolor}

\usepackage{wrapfig}
\usepackage{hyperref}
 \hypersetup{
     colorlinks=true,
     linkcolor=orange,
     filecolor=orange,
     citecolor=orange,      
     urlcolor=orange,
     }

\newcommand{\p}[1]{\noindent \textbf{{#1}.}}
\newcommand{\eq}[1]{Equation~(\ref{eq:#1})}
\newcommand{\fig}[1]{Figure~\ref{fig:#1}}

\title{TacStyle: Personalizing Tactile Robot Policies \\ using Structured Behavior Representations}

\author{
  Kevin Robledo*\\
  \And
  Matías I. Torres Galaz*\\
  \And
  Kumar Dixhant Rai \\
  \And
  Shelly Sara Ulman \\
  \And
  Tasmia Tasrin\\
  Department of Computer Science\\
  California State University, Northridge\\
  \texttt{tasmia.tasrin@csun.edu}\\
  \And
  Heramb Nemlekar\\
  Department of Mechanical Engineering\\
  California State University, Northridge\\
  \texttt{heramb.nemlekar@csun.edu}\\
}

\begin{document}
\maketitle

%===============================================================================
\vspace{-1ex}
\begin{abstract}
% Presonalization is needed
% process must be easy and efficient
% language is most natural
% labels are available for general tasks
% but labels difficult to describe fine adjustments to behavior
% Instead of directly conditioning robot behavior on language,
% We propose learning a structured representation of robot behaviors
% Enables foundational agents to reason over this space
% We show that much accurate and finer control
Robotic systems that assist humans should be capable of adapting their behaviors to individual user preferences. For instance, users may want a robotic arm to adjust the amount of force it applies while folding their laundry or cleaning furniture. Natural language provides an intuitive way for humans to communicate such preferences.
Recent progress in large-scale language-conditioned policies has shown that robots can successfully use language prompts to determine \emph{what} task to perform. However, extending the same approach to realize \emph{how} the task should be performed requires detailed labels describing the preferences or styles of trajectories in the task data.
Not only is collecting such annotations challenging, but conditioning directly on these labels may also fail to provide fine-grained control over 
a continuous range of behaviors.
For example, it can be difficult to convey the exact force that a robot must apply through abstract instructions like ``apply a bit more pressure than before''.
Therefore, in this work, we propose using language to reason over preferred behaviors instead of directly generating them.
We first learn a structured latent representation that organizes user preferences according to differences in the corresponding trajectories.
Then, given a preference prompt, we use a foundation model to interpret this latent space and choose a value that produces the desired behavior. Through both simulation and real-world experiments, we show that selecting robot behaviors from an intuitively structured latent space enables more precise adaptation to user preferences while requiring significantly fewer preference labels than language-conditioned policies.
% without any supervision. 
\end{abstract}

% Two or three meaningful keywords should be added here

%===============================================================================

\section{Introduction}

% PREVIOUS IDEA
% Recent efforts in collecting large, open-source datasets of robotic tasks~\cite{o2024open} have led to rapid progress in training generalist robot policies~\cite{zitkovich2023rt}.
% These datasets typically include trajectories demonstrating how a robot can accomplish everyday skills such as grasping objects, opening doors, or using hand-held tools.
% Using these datasets, robots can learn to imitate the primitive skills and even piece them together to perform complex, time-extended tasks~\cite{}.
% Most existing models rely on visual, natural language, or proprioceptive (e.g., encoder) sensory inputs during training and execution, which are sufficient for coarse, goal-directed motions. However, these modalities alone prove to be inadequate for fine, contact-rich manipulation tasks --- such as cleaning dishes or polishing furniture --- where precise force control and tactile feedback are crucial.

% NEW IDEA

Recent efforts in collecting large, open-source datasets of robotic tasks~\citep{o2024open} have led to rapid progress in training generalist robot policies~\citep{intelligence2026pi}. When these robots are eventually deployed in human settings, they will need to learn and adapt to individual preferences. 
For example, a robot assisting with physical rehabilitation may need to adjust the amount of force it applies when guiding users: some may want gentler assistance that gives them more control, while others may prefer a firmer support that feels more stable~\cite{molle2025exploring}. 
% Moreover, these preferences may also change over time. 

The problem of adapting robot policies has long been studied in the context of preference learning; much of the earlier work focused on inferring the parameters of the robot's reward function based on demonstrations of preferred trajectories~\cite{nemlekar2023transfer} and user responses to preference queries such as ``Do you prefer trajectory A or B?''~\cite{biyik2022learning}. 
However, these approaches either require significant human effort or many rounds of interaction which makes them less practical for everyday users. We therefore revisit this problem and ask: \textit{how can robots adapt to human preferences in a way that is both faster and more intuitive for users?}
% In this work, we revisit this problem and ask: what does preference adaptation look like in the era of pretrained robot policies?

% While language can be used to elicit preference queries, this underutilizes their reasoning capacity.
The advent of pre-trained language models (LMs) and vision-language models (VLMs) has made it much more convenient to use natural language to shape robot behavior.
Recent approaches either condition vision-language-action (VLA) policies on language embeddings~\cite{zitkovich2023rt} or use VLMs to steer robot actions~\cite{smith2025steer}.
Although these approaches have been successful at enabling discrete behavioral adjustments, such as switching tasks or specifying different target objects, extending the same paradigm to continuous and fine-grained behavioral preferences, such as controlling the amount of force a robot applies, would require users to provide large amounts of preference labels.

\begin{figure}[t]
    \begin{center}
        \includegraphics[width=0.99\textwidth]{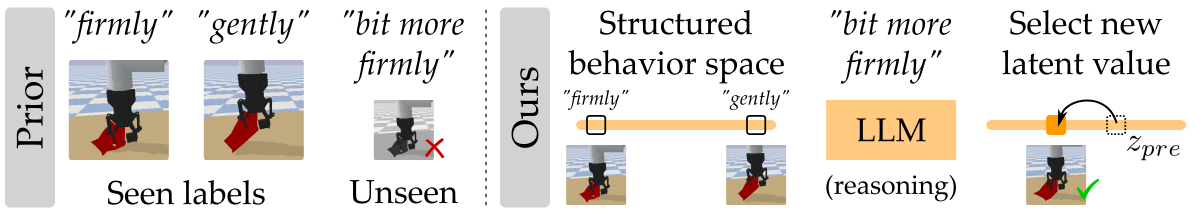}
        \vspace{-1 ex}
        \caption{Personalizing a robot arm to wipe with different levels of pressure. Robots that are trained with preference labels such as ``firm'' and ``gentle'', which only provide coarse descriptions of wiping styles, can fail to precisely interpolate to intermediate styles such as `` a bit more firmly''. For instance, the robot may default to a fully firm wipe.
        Instead, we encode wiping styles into a low-dimensional latent space and structure them proportionally to differences in behavior. This enables LLMs trained on internet-scale data to interpret the latent space and select latent values that produce intermediate wiping pressures, given only coarse labels for two opposite styles.}
        \label{fig:front}
        \vspace{-2 ex}
    \end{center}
\end{figure}

We hypothesize that this need for language-labeled data stems from the complexity of the representations learned by language-conditioned robot models, which makes it difficult to efficiently associate language prompts with corresponding robot behaviors (see \fig{front}).
Our key insight is that:
\begin{center}
\textit{Learning compact and structured robot behavior spaces enables LLMs to \\
naturally reason over continuous preferences without requiring extensive labels.}
\end{center}

Specifically, we propose encoding user preferences into a low-dimensional latent representation in which distances correspond linearly to differences in the associated robot trajectories. This structured representation is easier for LLMs to interpret and reason over. At runtime, users express their preferences through natural language, and the LLM selects a corresponding latent value, which is then provided as an input to the robot policy to produce the desired behavior. 
For example, a prompt such as ``be more gentle'' would map to a latent value associated with reduced force application. 
By reasoning over a structured behavior space rather than directly conditioning policies on preference prompts, our approach more effectively combines the reasoning capabilities of LLMs with the generality of pretrained robot policies.

Overall, our work presents a new perspective on how language can be leveraged to personalize robot behaviors more intuitively and precisely, and makes the following contributions:

\p{Learning structured latent representations of robot behavior}
We introduce an unsupervised framework for learning low-dimensional latent spaces in which distances are proportional to differences in user preferences, making it easier to map language prompts to corresponding behaviors.

\p{Leveraging LLMs for preference-guided behavior selection}
We show how LLMs can navigate the learned latent spaces with only sparse language supervision to both select latent values corresponding to user-specified preferences, like ``wipe firmly'', and progressively adjust them through comparative prompts such as ``apply a bit more pressure than before''.

\p{Evaluating against baselines in tactile tasks}
We evaluate our approach on both simulated and real-world tactile manipulation tasks, which remain underrepresented in large-scale robotic datasets and involve subtle variations that are difficult to describe through language.
We demonstrate that our method requires fewer preference-labeled language annotations while achieving the desired behaviors more precisely than recent vision-tactile-language-action (VTLA) models.

% proposed approach
% contributions

\section{Related Work}
% Para 1: brief talk about preference learning
% - what does it cover
% - how it has been done
% - what it essentially means, having the ability to adapt

% reward function / policy parameters
% different forms of feedback
% demonstrations and corrections need effort
% pairwise comparisons are easy to provide, but inference takes time
% more recently intantaenous approaches have been proposed such interface based selection or using natual language
% Interface learning can take time, 
% therefore we focus on natual language

% modalities of preference learning
% - language is easy
% - how is language used currently

\p{Preference Learning}
Humans often have different preferences for how their robotic systems should behave. Consider autonomous driving: some passengers may feel more comfortable with a cautious car that stays in the slow lane, while some may prefer a car that drives fast to save time. 
In such settings, preferences represent the distinct behavioral policies or driving styles~\citep{rosbach2019driving}.
Prior work has focused on learning such styles by adapting the robot’s reward function~\citep{losey2022physical} or policy~\citep{an2023direct} based on different forms of human feedback~\cite{mehta2023unified}, including trajectory demonstrations~\citep{ravichandar2020recent}, corrections~\citep{spencer2022expert}, and pairwise comparisons~\citep{biyik2022learning}.

%A key challenge in this problem is balancing the effort required from the user with how quickly and accurately the desired behavior can be achieved.
A key challenge in this problem is balancing the effort required from the user with the speed and accuracy of achieving the desired behavior.
Demonstrations and corrections can communicate rich preferences, but they often demand significant user effort and expertise. Pairwise comparisons are generally easier to provide, as users only need to choose between alternative trajectories; however, learning the underlying preference model can still require many rounds of interaction.
%since users only need to choose between alternative trajectories, but learning the underlying preference model can still require many rounds of interaction.
More recently, alternative modalities such as interface-based selection~\citep{nemlekar2025pecan} and natural language prompts~\citep{wu2023tidybot} have been explored. 
While interface-based methods enable precise control of robot parameters, natural language provides a more intuitive way for users to express preferences.
% Among these, interface-based approaches can still be cumbersome in practice, as users must first learn how to interpret and interact with the graphical interface.
We therefore focus on how language can be utilized for precise preference adaptation.

\p{Learning from Language Inputs}
Existing works have primarily used natural language in two ways: to condition robot policies and to enable high-level reasoning.
The most common approach is to condition a vision-language-action (VLA) policy on language inputs~\cite{ zitkovich2023rt, intelligence2026pi}.
These inputs may consist of direct language embeddings, shared vision-language representations learned using contrastive methods such as CLIP~\cite{radford2021learning}, or language representations aligned with encoded demonstration sequences~\citep{yu2023using,mees2022matters}.
Once these representations are learned, we can condition the robot's policy by either concatenating them with visual inputs or through mechanisms such as FiLM~\citep{perez2018film}.
In addition to these methods for generating low-level robot actions, we can use LLMs to extract behaviorally relevant features~\citep{peng2024preference} and decompose tasks into executable subtasks or motor primitives~\citep{ha2023scaling}. 

While there are some methods that leverage natural language to correct the robot's behavior during execution~\citep{cui2023no}, most efforts utilize language prompts to specify \emph{what} task the robot should perform, e.g., ``pick up the bottle''.
In contrast, our work utilizes language to specify \emph{how} the robot should perform the task, e.g., ``pick the bottle \textit{gently}''.
%we use language to specify \emph{how} the robot should perform the task, e.g., ``pick the bottle \textit{gently}''.

Only a small number of works have explored language-guided personalization. For instance, LLMs have been used to infer user preferences for object placement~\citep{wu2023tidybot} and to construct and select preferred decision trees~\citep{shankar2026low}.
However, these settings typically involve a small set of discrete choices, such as selecting between locations like `drawer' or `closet'.
% or choosing between behaviors such as `slow' and `fast'. 
On the contrary, we study continuous behavioral preferences in tactile manipulation tasks, such as controlling the amount of pressure applied while wiping a surface. Such preferences require fine-grained control of robot actions, making them substantially more challenging to specify and adapt to.

\p{Incorporating Tactile Inputs}
Recent works on contact-rich manipulation have similarly extended vision-language models to incorporate tactile feedback. 
Tactile embeddings can be directly concatenated with vision and language tokens to condition robot policies~\citep{zhang2026vtla}, or first aligned with the vision and language representations using contrastive learning~\citep{cheng2025omnivtla}, or fused with other modalities through Mixture-of-Experts (MoE) layers~\citep{yu2026forcevla}. 
% Shared representations can be learned between visual and tactile inputs~\citep{george2025vital}.
Descriptions of tactile observations can also be used to generate language prompts that are provided as inputs to a downstream VLA policy~\citep{bi2026vla}.

Regardless of how tactile feedback is incorporated, current approaches still primarily use language to specify discrete tasks rather than continuous human preferences. 
While the same policies could, in principle, be conditioned directly on preference descriptions, doing so would require large amounts of labeled data across different preference settings.
Instead, we aim to develop a more efficient way of leveraging language by enabling foundation models to reason over behavior representations.

\section{Problem Formulation}\label{sec:problem}

In this section, we define the components of the preference learning framework, including the robot states, actions, policies, and the data available for personalizing robot behavior.

\p{States and Actions}
Let $s_{t} \in \mathcal{S}$ be a tuple of all sensory inputs that the robot receives at any given time $t$.
For example, the states $s_{t}$ in our experiments include the robot's joint angles $x_{t}$, camera images $I_{t}$, and readings $\tau_{t}$ from a tactile sensor, i.e., $s_{t} = (x_{t}, I_{t}, \tau_{t})$.
Let $a_{t} \in \mathcal{A}$ denote the action taken by the robot at time $t$. In our experiments, actions correspond to the robot's joint velocities $\dot{x}_{t}$.

\p{Tasks and Styles}
We distinguish between tasks $\mathcal{G}$, which define \emph{what} the robot should do (e.g., pushing a door), and styles $\theta \in \Theta$, which specify \emph{how} the task should be performed depending on user preferences (e.g., slowly or forcefully).
%  qualitative descriptors

% \p{Generalist policy} We assume that the robot has access to a pre-trained generalist policy $\pi_{gen}$ that takes the states $s_{t}$ and language prompts $L$ as input, and outputs the actions $a_{t}$ the robot should take. In our experiments, actions $a_{t}$ correspond to the robot's joint velocities $\dot{x}_{t}$.
% $$a_{t} \leftarrow \pi_{gen}(s_{t}, L)$$

% Without loss of generality, we assume that the pre-trained policy $\pi_{gen}$ can complete the tasks described in the prompt, but only in a fixed, generic way. 
% For example, the policy may be able to push a door open, but only with a single level of force, without the ability to adjust its actions to be more gentle or more forceful as specified.
% In other words, it does not capture the different styles in which the same task can be performed.

\p{Policy}
We want the robot to not only accomplish the given task but also execute it according to the user’s preferred style.
Hence, the policy $\pi(a_{t} | s_{t}, \theta)$ must determine the robot action $a_{t}$ based on the task state $s_{t}$ at time $t$ as well as the desired style $\theta$. 
Since the user preferences $\theta$ are not directly known to the robot, the policy must instead learn them from an offline dataset of demonstrations $\mathcal{D}$ and adapt to them using language prompts $L$ provided by users at test time.
 
 \p{Dataset} 
We assume access to a dataset $\mathcal{D} = \{(\xi_{i}, P_{i})\}_{i=1}^N$ of trajectory and preference-label pairs. 
Each trajectory $\xi_{i} = \{(s_t, a_t)\}_{t=0}^{T}$ is a sequence of state-action pairs, and the associated label $P_{i}$ provides an absolute description of its style. 
These labels are non-comparative. They describe each trajectory independently, without specifying the relative differences between the trajectories. For example, multiple trajectories may be labeled as ``closing the door forcefully'', even though some trajectories may apply a bit more force than others.

We emphasize that our approach only requires preference labels for any two opposite styles in the dataset, $P_{max}$ and $P_{min}$, corresponding to trajectories that exhibit the minimum and maximum values of the underlying preferences (e.g., the gentlest and most forceful trajectories).
All trajectories with intermediate styles remain unlabeled when training our approach (i.e., $P_{i} = \emptyset$) and their labels are only used for training language-conditioned baselines.

\p{Language Prompts}
The language prompts $L$ are instructions provided by users at test time to specify their preferred execution style. Similar to the preference labels, these prompts may convey absolute preferences, such as ``close gently'', or unlike them, they may describe relative preferences, such as ``more gently than before.''

Our goal is to leverage these prompts to adapt the robot's actions such that its resulting trajectory reflects the user’s underlying preferred style $\theta$.

% ====================
% OTHER / OLD IDEAS

% Which tactile input is the most informative?

% How does tactile information affect manipulation performance? Which tasks does it affect the most?

% How best to combine tactile data with other modalities like vision and language?

% How can our data/sensors be combined with existing non-tactile datasets?

% Can we think of a general framework for integrating any new data modality with existing large datasets?

\section{Method}

We propose a new approach for generating preference-conditioned robot actions. 
Our method consists of two phases: the offline phase and the online phase. 
In the offline phase, we learn a latent representation of behavioral styles and train the robot policy to predict actions conditioned on these latent values. 
In the online phase, we leverage a large language model (LLM) to infer the appropriate latent styles from users’ natural language feedback. Our key insight is to structure this latent representation so that the LLM can accurately infer fine-grained style changes at test time.
We describe each component of this framework below.

\subsection{Learning Latent Styles Offline}

% 1 ------------
% To capture user preference, a style encode that takes entire trajectory and outputs a low-dimensional z value

% policy that takes a sequence of inputs and the style token and outputs actions

% training end-to-end ensures style will produce robot behavior

As mentioned in Section~\ref{sec:problem}, we assume access to a dataset of trajectories, each corresponding to some underlying preference or style $\theta$. However, these styles are not known to the robot \textit{a priori}. We therefore train the robot to learn these styles during training, based on how they influence the subsequent robot actions. Our architecture consists of two parts: a style encoder $f_{\phi}$ and a policy $\pi_{\theta}$.

\p{Style Encoder}
The style encoder takes each trajectory $\xi \in \mathcal{D}$ as input and maps it to a latent style value $z \in \mathbb{R}^{n}$. This latent value represents the trajectory's style.
$$f_{\phi}: \xi \mapsto z$$

\p{Policy Network}
The policy predicts the robot's actions $a_{t}$ at any step $t$ of the task. These actions depend on both the current state $s_{t}$ and the user's preferred style $\theta$. Since the true preference $\theta$ is not directly known, we instead condition the policy on the latent style $z$ output by the style encoder.
$$\pi_{\theta}: (s_{t}, z) \mapsto a_{t}$$

\begin{figure*}[t]
    \begin{center}
        \includegraphics[width=0.99\textwidth]{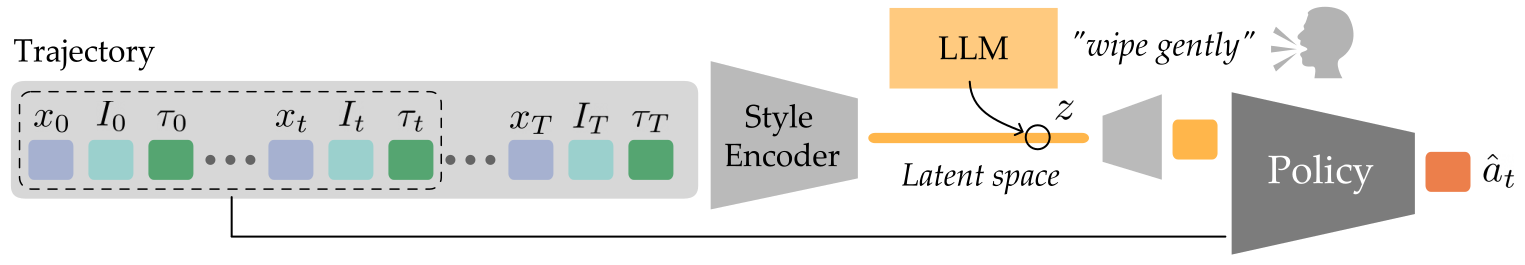}
        \vspace{-1 ex}
        \caption{Proposed architecture for personalizing robot behaviors. During training, the style encoder maps trajectories to points in a low-dimensional latent space. These latent values are then projected to the desired token size and provided as inputs to the policy network along with the task states to predict robot actions. We train the networks using the objective in \eq{combined_loss}.
        At test time, we replace the encoder with an LLM, which selects appropriate latent values based on user prompts.}
        \label{fig:model}
        \vspace{-2 ex}
    \end{center}
\end{figure*}

\fig{model} shows how the encoder and policy networks are connected. 
In our implementation, the policy receives a history of past states along with the latent value. 
We train the encoder and policy end-to-end by minimizing a behavior cloning loss to predict the actions demonstrated in the dataset.
\begin{equation}
    \mathcal{L}_{bc}(\theta, \phi) = - \mathbb{E}_{\xi \in \mathcal{D}} \left[ \mathbb{E}_{(s,a) \in \xi} \log \pi_{\theta} \left( a \mid s, f_{\phi}(\xi) \right) \right]
\end{equation}
% z value captures preference
Note that the expectation is taken over all state-action pairs of the trajectories in the dataset. 
To minimize this loss and accurately predict the actions, the policy must know the underlying style, as trajectories corresponding to different preferences can contain different actions for the same states. Thus, minimizing $\mathcal{L}_{bc}$ also encourages the encoder to learn latent representations that capture preference-relevant style information.

% MAYBE A PROOF CAN BE ADDED HERE, GIVEN THE ASSUMPTION THAT THE DATA CONTAINS DIFFERENT BEHAVIORS

\subsection{Mapping Natural Language to Latent Styles Online}\label{sec:method_llm}
% 2 -------------
% trajectory not available at test time, so z value cannnot be computed
% We want to determine z based on language prompt
% Specifically we want the provide an LLM with the entire Z space, few labels for points in the latent space, and the user prompt and ask it to decide a z value
% Prompt provided in appendix

During training, the latent style $z$ is obtained by encoding trajectories that demonstrate the underlying preference. 
At test time, however, asking users to provide demonstrations simply to communicate their preferred style is impractical.
Instead, we aim to enable users to specify preferences through natural language instructions $L$, such as “wipe the table with more pressure.” To achieve this, we need to map language descriptions of preferences to latent values that the policy can use.
%Instead, we want to enable users to specify preferences through natural language instructions $L$, such as ``wipe the table with more pressure''.
%To do so, we would need to map language descriptions of preferences to latent values that the policy can use.

Here, we use a foundation model $\mathcal{M}$, such as GPT-4~\cite{achiam2023gpt}, to perform this mapping.
$$\mathcal{M}: (z_{min}, z_{max}, P_{min}, P_{max},z_{pre}) \mapsto z_{new}$$
Given a user prompt $L$ and the robot's previous latent style $z_{pre}$ (if available), the model selects a new latent value $z_{new}$ from a continuous latent range $Z = [z_{min}, z_{max}]$. 
To guide this process, we provide it with preference labels corresponding only to the minimum ($z_{min}$) and maximum ($z_{max}$) latent values learned during training.
An example of this prompt is provided in Appendix~\ref{app:prompt}. 
% $Z = \{f_{\phi}(\xi) \mid \forall \xi \in \mathcal{D}\}$

This information alone may not be sufficient for the LLM to reliably select an appropriate latent value.
% While latent spaces learned through standard representation learning objectives often exhibit some degree of monotonicity~\cite{}, there is no guarantee that changes in latent values correspond proportionally to changes in robot behavior. 
For the LLM to meaningfully reason over the latent range $Z$, changes in latent values should lead to predictable changes in the robot’s style. 
Rather than providing extensive descriptions of this relationship, we propose structuring the latent representation such that the LLM can naturally infer and interpolate between styles with no additional explanations.

\subsection{Structuring Latent Style Representation}

We recommend two properties to make the latent space intuitive for an LLM to interpret. 
First, we constrain the latent space to be low-dimensional, such that $n << m$, where $m$ is the dimensionality of the state space $\mathcal{S}$. 
Second, we structure the representation to be linear such that the differences in latent values are proportional to differences in the corresponding trajectories:
\begin{equation}
    |z_{i} - z_{j}| \propto \beta \; d(\xi_{i}, \xi_{j})
\end{equation}
where $d(\xi_{i}, \xi_{j})$ measures the behavioral difference between two trajectories and $\beta$ is the proportionality constant.
In our experiments, we approximate $d(\cdot)$ using the average Euclidean distance between time-aligned states from the trajectory pairs.

To enforce this structure, we introduce the following regularization objective:
\begin{equation}
    \mathcal{L}_{prop}(\phi, \beta) = \mathbb{E}_{(\xi_{i},\xi_{j \neq i}) \in \mathcal{D}} \left[ \left( |z_i - z_j| - \beta \, d(\xi_{i}, \xi_{j}) \right)^{2} \right]
\end{equation}
Minimizing $\mathcal{L}_{prop}$ encourages nearby latent values to map to similar robot behaviors, while larger differences in the latent space correspond to proportionally larger behavioral changes.

Therefore, during training, we optimize the combined objective to train the encoder and policy:
\begin{equation}
    \mathcal{L}(\theta, \phi, \beta) = \mathcal{L}_{bc} + \lambda \, \mathcal{L}_{prop} \label{eq:combined_loss}
\end{equation}
$\lambda$ controls the trade-off between accurately imitating the demonstrated actions and enforcing structure in the latent space. Together, these objectives enable the latent representation to capture relevant style information while being interpretable enough for LLMs to make predictable style adjustments.

% PROVIDE EXPLANATION HERE THAT 
% prompt --> latent value (is what we make linear)
% Latent value --> action (depends on data and ability of policy to generalize)
% Some justification is needed here as to why linear mapping to latent value helps
% i.e. how does linearization of inputs helps produce desired behavior more accurately

% \subsection{Implementation}
We provide further implementation details in Appendix~\ref{app:implement} and our code can be accessed at: \url{https://github.com/Anonymouslab2026/TacStyle}

\section{Experiments}\label{sec:experiments}

% OLD NOTES:
% We consider the following tasks for evaluating our proposed approach:
% \begin{enumerate}
%     \item Grasping a sponge
%     \item Wiping a plate
%     \item Placing a plate in the dishwasher
%     \item Closing the dishwasher
% \end{enumerate}
% The first three tasks represent canonical scenarios in which tactile feedback can play a vital role. While versions of these tasks have been explored in prior work, a key distinction of our evaluation is that we do not rely on simplified, proof-of-concept instances (e.g., performing a single wiping motion across the plate). Instead, we consider realistic, complete implementations of these activities, for example, requiring multiple wiping motions at varying angles and contact pressures to clean the plate.
% The last task of closing the dishwasher serves as a control scenario in which tactile sensing may not be as critical.

%% NEW:

In this section, we evaluate whether using language to select preferred behaviors from a structured representation is more effective than directly conditioning robot actions on language. 
We hypothesize that leveraging LLMs to infer styles from a proportional latent space will produce trajectories that better match users’ true preferences compared to recent language-conditioned baselines.

% Add more detail of the simulation.  

% information about the states, images, tactile sensor in each task

% description of styles in each task 

%insert image of green shirt with tape to show pickup locations ^^^
% information about the data collected in each task

\p{Experimental Setup} 
We test our approach in two environments: a simulated wiping task in PyBullet~\cite{coumans2021} and a real-world cloth-folding task using a Fairino FR5 robotic arm.

In the simulated task, a 6-DoF robotic arm grasped a deformable cloth and wiped it across a table in a straight line.
The underlying preference in this task corresponded to the height at which the robot moved, which directly affected the amount of pressure applied to the table. 
To train the robot, we collected $30$ demonstrations spanning $5$ wiping heights. At each time step, we recorded the robot's $6$-D joint angles, RGB images from a third-person camera, $1$-D gripper state, $3$-D tactile readings, and $7$-D actions consisting of $6$-D joint velocities and $1$-D gripper command. 
We obtained the tactile readings using a simulated DIGIT sensor~\cite{lambeta2020digit} implemented in Tactile Gym~\cite{lin2022tactilegym2}.

For the real-world experiments, a FR5 Fairino robotic arm collaboratively folded a t-shirt with a human operator by grasping opposite shoulder ends of the cloth.
The folding style ranged from loose (cloth sagging) to tight (cloth stretched), depending on the robot’s grasping location.
We collected $15$ demonstrations spanning $3$ different tightness configurations.
At each time step, we recorded the robot's $6$-D joints, $2$-D gripper state, $15$-D tactile readings, and $89$-D actions.
We obtained the tactile readings from a custom-fabricated eFlesh sensor~\cite{pattabiraman2025eflesh}.

\begin{figure*}[t]
    \centering
    \includegraphics[width=\textwidth]{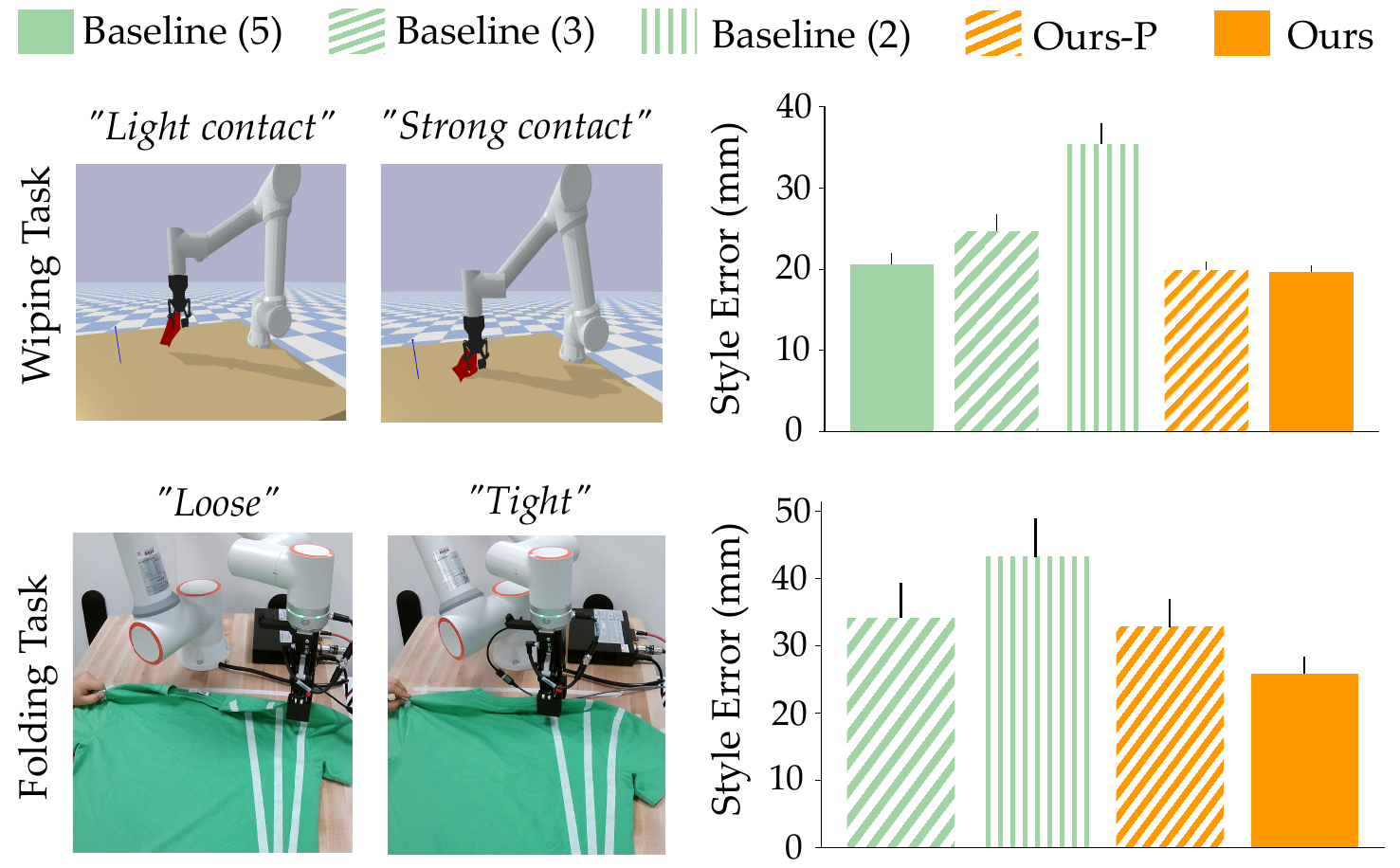}
    \vspace{-2ex}
    \caption{Comparison of methods for adapting robot behaviors based on language prompts. The top row shows results for the simulated wiping task, where styles range from light to strong contact, while the bottom row shows results for the real-world folding task, where styles range from loose to tight folds. 
    We see that our proposed approach, which leverages structured style representations, achieves the target styles more accurately than language-conditioned baselines.
    Only \textit{Baseline (5)} performs comparably to \textit{Ours} in the simulated task, but it requires preference labels for all styles in the training data, whereas our approach uses labels for only the two extreme styles.
    % Lower style error indicates better agreement between the generated behavior and the desired contact style.
    The error bars indicate standard error.}
    \label{fig:results}
    \vspace{-2ex}
\end{figure*}

% description of baseline implementation
% versions of baseline

\p{Independent Variables}
In the simulation experiments, we evaluate our approach (\textit{Ours}) against a Vision-Tactile-Language-Action (VTLA) baseline~\cite{zhang2026vtla}. We implement the baseline using a pretrained Qwen vision-language model~\cite{Qwen2-VL} to encode language prompts, which are then provided as inputs to a transformer-based policy. For consistency, both the baseline and our approach use the same transformer policy architecture, input token dimensionality, and conditioning mechanism.
For our method, we first encode demonstrations into a low-dimensional latent space ($n=1$). We then use GPT-4o mini to infer latent values from user prompts, which are projected back to the policy token dimensionality.
For the real-world experiments, we use the same setup with one modification: since RGB observations are not recorded, we compare against a tactile-language-action (TLA) baseline~\cite{hao2026tla} instead of a vision-language baseline.
The preference labels for training the baselines and the language prompts for personalizing the robot during evaluation are generated using a separate GPT-4o mini instance that is provided with additional domain knowledge to act as an oracle human prompter. Examples of the generated prompts are described in Appendix~\ref{app:prompt}.

In total, we compare \textit{Ours} with four methods: three versions of the baseline trained with preference labels for different numbers of styles in the training data, namely \textit{Baseline (5)}, \textit{Baseline (3)}, and \textit{Baseline (2)}, and an ablation of our method trained without $\mathcal{L}_{prop}$ loss (\textit{Ours-P}).
For example, when training \textit{Baseline (2)}, we provide preference labels for all trajectories corresponding to only $2$ out of the $5$ styles in the wiping task.
In contrast, both versions of our approach require labels for only two trajectories in the entire dataset.

\p{Dependent Variables}
Our goal is to assess how effectively our approach meets the desired preferences.
%Our goal is to evaluate how well our approach achieves the desired preferences. 
For this, we first uniformly sample 7 target wiping styles and 5 target folding styles, including intermediate styles that were not seen during training.
We then use the oracle GPT instance to generate natural language prompts corresponding to each target style. Given these prompts, we roll out trajectories for each method and measure the mean absolute difference between the target style and the style achieved by the resulting trajectory.
We use the distance between the desired and rolled out end-effector positions as a measure of the \textit{Style Error}.

\p{Results}
The results averaged over $10$ training and testing runs for the simulated task and $5$ runs for the real-world task are shown in \fig{results}.
Our approach (\textit{Ours}) of selecting styles from a proportional latent representation achieves the lowest style error across both tasks. While the ablation without the proportionality loss (\textit{Ours-P}) performs similarly to \textit{Ours} in the wiping task, where a small number of joints primarily control differences in styles, incorporating the proportionality loss substantially improves the accuracy of the resulting behaviors in the more complex folding task.
% where styles involve larger variations in joints.
We also observe that the language-conditioned baselines, specifically \textit{Baseline (5)} for wiping and \textit{Baseline (3)} for folding, perform comparably to our approach when trained with preference labels for all styles present in the data. However, the baseline performance degrades significantly as the number of preference labels is decreased. In particular, \textit{Baseline (2)}, which receives preference labels only for the two extreme styles (the same available to the LLM in \textit{Ours}), performs the worst across both tasks. 
Overall, these results highlight the advantage of reasoning over structured latent spaces rather than directly conditioning policies on language.
In Appendix~\ref{app:relative}, we present additional experiments with relative prompts (e.g., ``fold more tightly than before''), where LLM reasoning over previous style selections ($z_{pre}$) allows our approach to progressively adapt the robot’s behavior, unlike language-conditioned baselines that only consider the latest prompt.

\section{Limitations}

% 1. Dimensionality of styles and limited style ranges
% one-dimensional styles, multi-dimensional styles need disentangelement
% limited style range,

\p{Multi-dimensional Styles}
A key limitation of our work is that we focus our analysis on learning one-dimensional style representations that capture a single underlying preference axis. While this assumption is a reasonable starting point, further work is needed to capture preferences that involve trade-offs between multiple preference axes, such as speed, safety, force, and efficiency. For example, users may say ``wipe firmly but slowly.'' Extending our framework to higher-dimensional latent spaces would likely require disentangling these preference axes~\cite{mathieu2019disentangling, wang2024disentangled} to ensure that foundational models can still reliably reason over the learned representations.

% 2. Conditioning approach 
% we condition by appending, FiLM for all approaches

% 3. userstudy
% use orcale LLM as human prompter, user study next
% tactile forces correlated to 

\p{Human Evaluations}
Another limitation is that we evaluate language interaction using prompts generated by an oracle LLM with additional domain knowledge, rather than through real user studies. While we introduce variability in the prompts generated by the LLM, real users may express preferences more ambiguously or with varying levels of specificity. 
A user study would also provide insights into how users perceive our system, beyond the objective performance metric of style error.

% 3. broader application
% test in only two tasks,
% single task models, multi-task models, 
% on same note, we train our own policies, but extend to pre-trained policies on large-scale

\p{Scaling to Large-Scale Robotic Systems}
Finally, our experiments focus on task-specific policies for two contact-rich manipulation tasks. While these tactile tasks serve as useful examples of settings with continuous preferences, an important open question remains regarding how the proposed latent structuring approach scales to broader multi-task robot models trained on large-scale datasets.
%While they are useful examples of settings with continuous preferences, an important open question is how the proposed latent structuring approach scales to broader multi-task robotic systems and pretrained VLA models trained on large-scale datasets.

\section{Conclusion}

% highlight that we focus on continous preferences

% Sure we can always collect more data, but imagine a deployed system, would we want end-users to provide this labels, giving users the power to simply annotate extreme styles and then being able to interpolate between 

%In this work, we studied how robots can adapt to continuous user preferences through natural language. Rather than directly conditioning robot policies on language prompts, we proposed learning a structured latent representation of robot behavior that enables LLMs to infer and interpolate between styles.
%Through both simulated and real-world tactile tasks, we showed that our approach improves the accuracy of preference adaptation while requiring significantly fewer language-labeled demonstrations than recent language-conditioned baselines.

In this work, we investigated how robots can adapt to continuous user preferences expressed through natural language. Instead of directly conditioning robot policies on language prompts, we proposed learning a structured latent representation of robot behavior that enables LLMs to infer and interpolate between behavioral styles. 
Through experiments on both simulated and real-world tactile manipulation tasks, we demonstrated that our approach improves the accuracy of preference adaptation while requiring significantly fewer language-annotated demonstrations compared to recent language-conditioned baselines.
Videos of our approach and experiments can be found at: \url{https://anonymouslab2026.github.io/TacStyle/}.

More broadly, our work presents an alternative perspective on personalization in robotics. Although it is always possible to collect larger amounts of preference-labeled data, doing so may be impractical for deployed robotic systems used by everyday users. Instead, our results suggest that users may only need to specify a small number of labels (e.g., the gentlest and forceful styles) to enable the system to achieve intermediate behaviors. We look forward to exploring how the proposed latent structuring approach can be extended to large-scale, multi-style, and multi-task settings.

%===============================================================================

\clearpage
% The acknowledgments are automatically included only in the final and preprint versions of the paper.
% \acknowledgments{If a paper is accepted, the final camera-ready version will (and probably should) include acknowledgments. All acknowledgments go at the end of the paper, including thanks to reviewers who gave useful comments, to colleagues who contributed to the ideas, and to funding agencies and corporate sponsors that provided financial support.}

%===============================================================================

% no \bibliographystyle is required, since the corl style is automatically used.
\bibliography{references}  % .bib
\clearpage

\appendix

\section{Appendix}

\subsection{Implementation} \label{app:implement}

For all methods in our experiments, visual observations were encoded using a ResNet-18 image encoder when RGB inputs were available. The resulting visual features, robot states, and tactile readings were then projected into token representations using multilayer perceptrons (MLPs). 
For our approach, token sequences from each demonstration were mapped to a one-dimensional latent representation using a gated recurrent unit (GRU)-based style encoder. 
For the language-conditioned baselines, instead of a style encoder, we used a frozen Qwen2 language encoder to convert preference labels into semantic embeddings, which were subsequently projected to the policy token
dimensionality using an MLP.
All networks were trained using the Adam optimizer with a learning rate of $1 \times 10^{-4}$. To ensure fair comparison, all methods were trained with the same number of demonstrations and the same batch size until they reached comparable training losses.

\subsection{Preference Labels and LLM Prompts} \label{app:prompt}

In our experiments, we used an LLM to generate the preference labels describing the style of each trajectory in the training data. This oracle LLM was instructed to respond like a human annotator and was provided with a description of the task as well as examples of labels for trajectories with different styles. Tables~\ref{tab:wiping_prompts} and~\ref{tab:folding_prompts} show examples of the natural language labels generated by the oracle in the wiping and folding tasks, respectively.

\begin{table*}[h]
\centering
\begin{tabular}{c p{0.82\linewidth}}
\hline
\textbf{Wiping Height} & \textbf{Preference Labels} \\
\hline

0.050 
% & ``wipe the cloth with firm pressure'' \\
& ``wipe the cloth firmly with low height'' \\
& ``wipe the cloth with strong pressure'' \\
& ``wipe the surface with firm pressure'' \\

\hline

0.075 
& ``wipe the surface with firm contact'' \\
% & ``wipe the cloth with firm contact'' \\
& ``wipe the cloth with firm contact and moderate height'' \\
& ``wipe the cloth with firm contact at a low height'' \\
% & ``wipe the cloth with firm contact at a lower height'' \\

\hline

0.100 
& ``wipe the cloth with moderate pressure'' \\
& ``wipe the surface with moderate contact'' \\
& ``perform a moderate contact wipe'' \\

\hline

0.125 
% & ``wipe with gentle contact at a moderate height'' \\
& ``wipe with a gentle touch and moderate contact'' \\
& ``wipe with a balanced contact force'' \\
& ``wipe with mild pressure and moderate height'' \\
& ``wipe softly with moderate contact'' \\

\hline

0.150 
% & ``wipe the cloth softly with a gentle touch'' \\
% & ``wipe softly with a higher height'' \\
& ``wipe softly with a higher height for lighter contact'' \\
% & ``gently wipe the cloth with a light touch'' \\
% & ``wipe softly with a light touch'' \\
& ``wipe softly with lighter pressure and increased height'' \\
& ``gently wipe the surface with a light touch'' \\
& ``wipe softly with a higher contact height'' \\
% & ``gently wipe the cloth with a lighter touch'' \\

\hline
\end{tabular}
\caption{Generated language prompts for simulated wiping styles, where lower wiping heights correspond to higher contact pressures and higher wiping heights correspond to lower contact pressures.}
\label{tab:wiping_prompts}
\end{table*}

\begin{table*}[h]
\centering
\begin{tabular}{c p{0.82\linewidth}}
\hline
\textbf{Fold Tightness} & \textbf{Preference Labels} \\
\hline

0.0 
& ``fold the cloth very loosely'' \\
& ``fold the cloth gently'' \\

\hline

0.5
& ``fold the cloth with a moderate tightness'' \\
& `` fold the cloth moderately'' \\

\hline

1.0 
& ``fold the cloth firmly'' \\

\hline
\end{tabular}
\caption{Generated language prompts for folding styles ranging from loose (0.0) to tight (1.0).}
\label{tab:folding_prompts}
\end{table*}

\begin{figure}[h]
    \centering
    \includegraphics[width=0.7\linewidth]{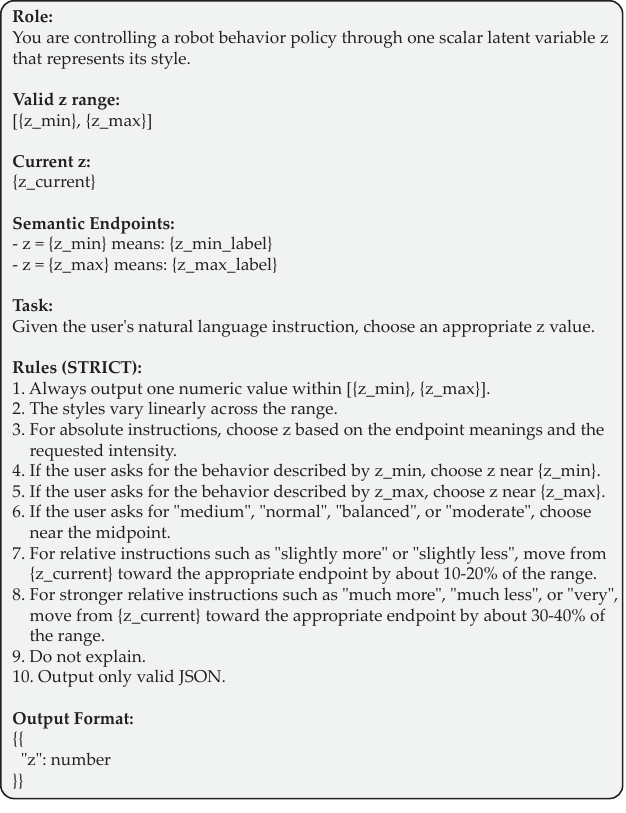}
    \caption{
    Prompt used to initialize the preference-reasoning LLM. Given the labels for two opposite styles and a user preference prompt, the LLM reasons over the structured latent space to select an appropriate latent style value. Here, \texttt{z\_current}, \texttt{z\_min\_label}, and \texttt{z\_max\_label} correspond to $z_{prev}$, $P_{min}$, and $P_{max}$, respectively, as defined in Section~\ref{sec:method_llm}.
    }
    \label{fig:system_prompt}
\end{figure}

When training our approach, we use preference labels only for the lowest and highest wiping heights or folding tightness levels. 
These two labels are then provided as reference points to the preference-reasoning LLM defined in Section~\ref{sec:method_llm}, which is different from the oracle LLM used to annotate the dataset. 
The system prompt used to initialize this LLM is shown in \fig{system_prompt}.

\subsection{Experiments with Relative Preference Prompts} \label{app:relative}

In Section~\ref{sec:experiments}, we evaluated personalization of robot behaviors based on natural language descriptions of absolute preferences such as ``fold the cloth tightly''. Here, we consider a more challenging setting in which users provide relative preference prompts such as ``fold a bit more tightly than before''.
We compare our full approach (\textit{Ours}) against the strongest language-conditioned baseline in the folding task, \textit{Baseline (3)}, which is trained with preference labels for all training demonstrations

Specifically, we initialize our approach with $z_{prev}$ corresponding to the tightest fold, and repeatedly provide the prompt ``fold a bit more loosely than before'' for five iterations. 
In each iteration, the preference-reasoning LLM selects a new latent style $z_{new}$ based on the prompt. We then roll out the policy using $z_{new}$ and record the style achieved by the resulting trajectory.
Before processing the next prompt, we update $z_{prev}$ to be the latent value selected in the previous iteration.
Since the language-conditioned baseline cannot directly condition on previously selected styles, we introduce a variant, \textit{Relative Baseline (3)}, in which an oracle LLM generates a natural language description of the previously achieved style and appends it to the prompt.
For example, the prompt is modified to include  ``... given that the previous style was \texttt{[previous style label]}.''

We repeat the same procedure starting from a $z_{prev}$ corresponding to the loosest fold, and repeatedly provide the prompt ``fold a bit more tightly than before'' for three iterations.

\begin{figure*}[t]
    \centering
    \includegraphics[width=\textwidth]{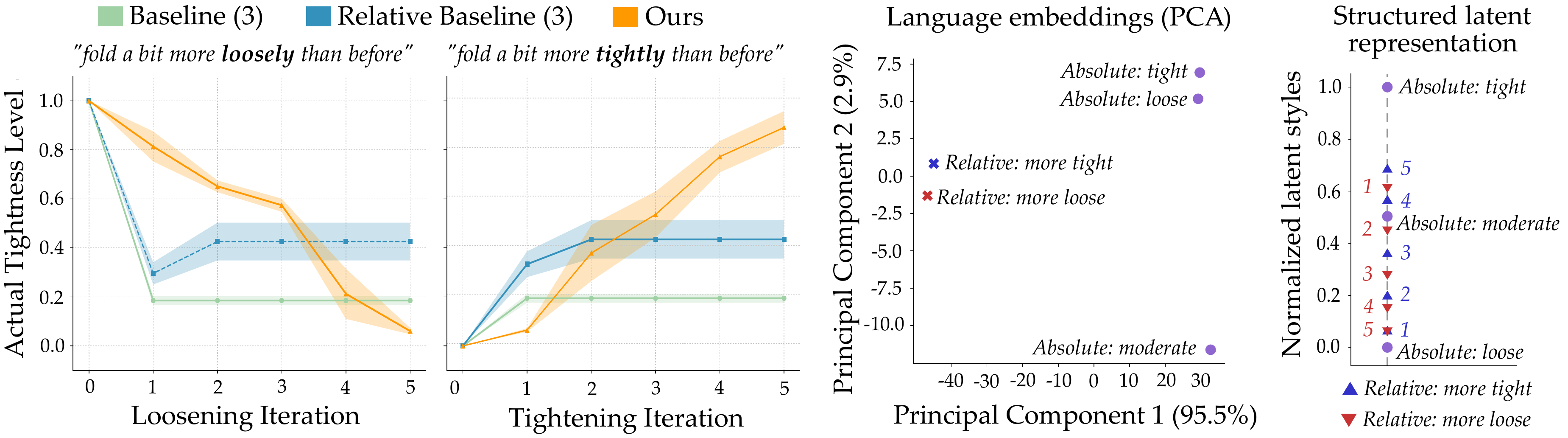}
    \vspace{-2ex}
    \caption{Relative preference adaptation in the folding task. The two plots on the left track the average tightness achieved after successive prompts specifying relative folding preferences: ``more tightly'' or ``more loosely''.
    While our approach progressively adapts the folding style in the desired direction, the language-conditioned baselines converge to similar behaviors in both cases. 
    The third plot visualizes the language embeddings used by \textit{Baseline (3)} for absolute and relative preference prompts using Principal Component Analysis (PCA). We see that the embeddings for tight, more tight, moderate, loose, and more loose queries are not organized intuitively. 
    The rightmost plot shows the normalized latent space learned by our approach, in which intermediate styles are arranged linearly between the tight and loose extremes. The error bands indicate standard error.}
    \label{fig:results_appendix}
    \vspace{-1ex}
\end{figure*}

\p{Results}
The left side of \fig{results_appendix} summarizes the average folding styles produced by trained models from the same $5$ runs as in Section~\ref{sec:experiments}.
We observe that our approach progressively adjusts the folding style, making it increasingly tighter or looser after each relative prompt.
In contrast, both baselines quickly converge to a fixed folding tightness and exhibit no incremental change in style thereafter.

The lack of incremental adaptation is expected because the baselines do not explicitly reason over previous styles.
However, to better understand why they converge to similar styles when given opposite requests of folding ``more tightly'' and ``more loosely'', we analyze the language embeddings they use to condition the policy.
Specifically, the right side of \fig{results_appendix} shows a plot of the first two principal components of the embeddings generated by the Qwen model. We observe no clear directional relationship between the relative prompts. Instead, the encoder maps these prompts to nearby semantic embeddings, causing the policy to produce similar behaviors in each case.
On the other hand, the latent values selected by our approach, as defined in Section~\ref{sec:method_llm}, change progressively in response to the relative prompts. 
These latent styles monotonically modify the robot's behavior in the requested direction, as shown in the plots on the left.

These results highlight another benefit of structuring latent spaces for incremental language-based preference adaptation. By learning representations that LLMs can interpret, our approach enables fine-grained and iterative personalization without requiring large amounts of preference-labeled data. While this work focuses on styles that can be represented along a single preference axis, we look forward to extending our framework to richer multi-dimensional preferences in future work.

\end{document}